\documentclass{article}
\usepackage{iclr2027_conference,times}
\usepackage{amsmath,amsfonts,amssymb,amsthm,bm}
\usepackage{algorithm}
\usepackage{algpseudocode}
\usepackage{booktabs}
\usepackage{array}
\usepackage{multirow}
\usepackage[olditem,oldenum]{paralist}
\usepackage{enumitem}
\usepackage{hyperref}
\usepackage{url}
\usepackage{bbding}
\usepackage{pifont}

\newtheorem{theorem}{Theorem}
\newtheorem{lemma}[theorem]{Lemma}
\newtheorem{assumption}{Assumption}
\newtheorem{definition}{Definition}

\newcommand{\X}{\mathcal{X}}
\newcommand{\Y}{\mathcal{Y}}
\newcommand{\vV}{\mathbf{V}}
\newcommand{\vdelta}{\boldsymbol{\delta}}
\newcommand{\ind}{\mathbf{1}}
\newcommand{\norm}[1]{\lVert #1\rVert}
\newcommand{\inner}[2]{\langle #1,#2\rangle}
\newcommand{\DReg}{\text{\normalfont D-Reg}}
\newcommand{\SW}{\text{\normalfont SW-Reg}}
\DeclareMathOperator{\diam}{diam}

\def\1{\bm{1}}

\def \rightsym  {\ding{51}}
\def \wrongsym  {\ding{55}}

\newcommand{\vh}{\mathbf{h}}

\newcommand{\vn}{\mathbf{n}}

\newcommand{\vu}{\mathbf{u}}
\newcommand{\vv}{\mathbf{v}}
\newcommand{\vw}{\mathbf{w}}
\newcommand{\vx}{\mathbf{x}}
\newcommand{\vy}{\mathbf{y}}
\def\vz{{\bm{z}}}

\DeclareMathAlphabet{\mathsfit}{\encodingdefault}{\sfdefault}{m}{sl}
\SetMathAlphabet{\mathsfit}{bold}{\encodingdefault}{\sfdefault}{bx}{n}

\def\gI{{\mathcal{I}}}

\def\gO{{\mathcal{O}}}

\def\gS{{\mathcal{S}}}

\newcommand{\E}{\mathbb{E}}

\newcommand{\R}{\mathbb{R}}

\DeclareMathOperator*{\argmin}{arg\,min}

\title{From Switching to Dynamic Regret: A Simple \\ Reduction via Unbiased Random Sequences}

\author{%
\textbf{Yibo Wang}\textnormal{\textsuperscript{1,2}},
\textbf{Wenhao Yang}\textnormal{\textsuperscript{1,2}},
\textbf{Sifan Yang}\textnormal{\textsuperscript{1,2}},
\textbf{Wei Jiang}\textnormal{\textsuperscript{3}},
\textbf{Yuanyu Wan}\textnormal{\textsuperscript{4}},
\textbf{Lijun Zhang}\textnormal{\textsuperscript{1,2,}}\thanks{  Corresponding author: zhanglj@lamda.nju.edu.cn} \\
\textsuperscript{1}State Key Laboratory for Novel Software Technology, Nanjing University \\
\textsuperscript{2}School of Artificial Intelligence, Nanjing University \\
\textsuperscript{3}School of Computer Science and Engineering, Nanjing University of Science and Technology \\
\textsuperscript{4}School of Software Technology, Zhejiang University
}

\iclrfinalcopy 

\begin{document}
\maketitle

\begin{abstract}
    In non-stationary online learning, dynamic regret has attracted increasing attention as a measure of how well an online learner performs against a time-varying comparator sequence. Despite considerable advances, existing methods to attain optimal dynamic regret bounds for strongly convex losses and exp-concave losses remain analytically intricate or computationally demanding. In this paper, we present a \textit{simple} framework that reduces dynamic regret minimization to switching regret minimization. The key idea is to construct, for any comparator sequence, an auxiliary random sequence that is unbiased at each round, with controlled variance and a manageable number of switches. Combining this construction with suitable surrogate losses, we decompose dynamic regret into the expected switching regret against the infrequently-changing random sequence and a variance-controlled approximation error. Consequently, \textit{any} algorithm with a switching regret guarantee can be plugged into our framework to obtain a corresponding dynamic regret bound. In particular, by instantiating the framework with existing switching-regret algorithms, we can obtain an $\widetilde{\gO}(T^{1/3}P_T^{2/3})$ dynamic regret bound for strongly convex losses and $\widetilde{\gO}(d^{2/3}T^{1/3}P_T^{2/3})$ for exp-concave losses, where $P_T$ denotes the path-length and $d$ denotes the dimension. For general convex losses, the same reduction recovers the $\gO(\sqrt{T(1+P_T)})$ dynamic regret bound. Notably, all our findings match the known minimax-optimal bounds, up to logarithmic factors, and the modularity of our framework makes it easy to understand and implement.
\end{abstract}
\section{Introduction}
\label{sec:introduction}

Non-stationary online learning investigates sequential decision-making in evolving environments, with many real-world applications such as online recommendation and traffic scheduling \citep{Others:2016:Hazan}. A standard framework for studying such problems is Online Convex Optimization (OCO) \citep{Others:2012:Shalev-Shwartz,ArXiv:2019:Orabona}, which can be viewed as a repeated game between a learner against the environment. Formally, at each round $t$, the learner chooses a decision $\vx_t$ from a convex set $\X$, and then suffers a loss $f_t(\vx_t)$ with the convex function $f_t(\cdot): \X \rightarrow \mathbb{R}$ revealed by the environment. The goal of the learner is to minimize the dynamic regret \citep{ICML:2003:Zinkevich}:
\begin{equation}
    \label{eq:dynamic_regret}
    \DReg_T(\vu_{1:T})=\sum_{t=1}^T f_t(\vx_t)- \sum_{t=1}^T f_t(\vu_t),
\end{equation}
which is defined as the difference between the cumulative loss incurred by the learner and that of \textit{any} possible comparators $\vu_1, \dots, \vu_T \in \X$.  By restricting the comparator sequence, \eqref{eq:dynamic_regret} recovers different metrics in OCO. For example, choosing the best fixed comparator, with $\vu_t=\vu \in \argmin_{\vx \in \X} \sum_{t=1}^T f_t(\vx)$ for all $t$, yields the static regret \citep{Others:2006:Bianchi}, and choosing the best piecewise-constant comparator sequence with a bounded number of switches delivers the switching regret \citep{NeurIPS:2024:Pasteris,ICML:2026:Yang}.

Over the years, there have been a variety of investigations into dynamic regret minimization owing to its generality \citep{ICML:2013:Hall,Others:2015:Besbes,AISTATS:2015:Jadbabaie,CDC:2016:Mokhtari,ICML:2016:Yang,NeurIPS:2016:Wei,ICML:2018:Zhang,NeurIPS:2019:Baby,AISTATS:2020:Zhang,ICML:2020:Cutkosky,AAAI:2021:Wan:B,NeurIPS:2021:Zhang:B,COLT:2023:Wan,AAAI:2024:Wang}. In particular, the seminal work of \citet{ICML:2003:Zinkevich} introduces the path length $P_T$, defined in \eqref{eq:pt}, to capture the fluctuations of the comparator sequence, and establishes an $\gO(\sqrt{T}(1+P_T))$ dynamic regret bound for general convex losses. Subsequently, \citet{NeurIPS:2018:Zhang} improve this result to the provably optimal $\gO(\sqrt{T(1+P_T)})$ bound. For strongly convex losses and exp-concave losses, \citet{COLT:2021:Baby} establish $\widetilde{\gO}(d^{7/3}T^{1/3}P_T^{2/3})$ and $\widetilde{\gO}(d^{23/6}T^{1/3}P_T^{2/3})$  bounds, respectively,  through delicate analyses of the KKT conditions on offline optimal comparator sequences. Using similar analyses, \citet{AISTATS:2022:Baby} reduce the dimension dependence  for strongly convex losses and exp-concave losses  to $d^{2/3}$ and $d^{10/3}$, respectively. Recently, \citet{ICML:2025:Zhang} further improve the dimension dependence for exp-concave losses, and obtain an $\widetilde{\gO}(dT^{1/3}P_T^{2/3})$ dynamic regret bound. However, their method requires Kullback-Leibler (KL) projections onto constrained Gaussian mixtures, incurring substantial computational costs and complicating implementation.

In this paper, we develop a \textit{simple} framework that reduces dynamic regret minimization to switching regret minimization, avoiding both intricate analyses of offline optimal comparator sequences and computationally intractable KL projections. Specifically, we first show that, for \textit{any} comparator sequence, one can construct an auxiliary random sequence that follows it in expectation with bounded variance and a controlled expected number of switches. Combined with suitable surrogate losses, these properties allow us to decompose dynamic regret to the expected switching regret against the infrequently-changing auxiliary sequence and a variance-controlled approximation error. Consequently, existing algorithms with switching regret guarantees can be used as black boxes to obtain dynamic regret bounds. Importantly, the auxiliary random sequence is introduced solely for the analysis and need not be explicitly generated by the learner. In practice, the learner only runs the chosen switching-regret algorithm on the surrogate losses, without any access to or operations on the comparator sequence. 
Our main contributions are summarized as follows.

\begin{compactitem}[\hspace{0.00cm}$\bullet$]

    \item  We show that, for any comparator sequence, one can construct an unbiased auxiliary random sequence with bounded variance and few switches in expectation, enabling a reduction from dynamic to switching regret with a variance-controlled approximation error.  
   
    \item Algorithmically, we develop a simple and modular framework that reuses existing switching-regret algorithms as black boxes, without modifying their internal updates.
    
    \item Theoretically, our framework yields dynamic regret bounds of $\widetilde{\gO}(T^{1/3}P_T^{2/3})$ for strongly convex losses, $\widetilde{\gO}(d^{2/3}T^{1/3}P_T^{2/3})$ for exp-concave losses, and $\gO(\sqrt{T(1+P_T)})$ for general convex losses. In particular, we improve the dimension dependence for strongly convex losses and exp-concave losses. All our results  attain the optimal dynamic regret bounds, up to logarithmic factors. 

\end{compactitem}   
\section{Related Work}
\label{sec:related_work}

\begin{table}[t]
\centering
\caption{Comparisons of methods for dynamic regret minimization. Abbreviations: cvx $\to$ convex, str-cvx $\to$ strongly convex, exp-concave $\to$ exponentially concave. The last column indicates whether the proposed method is computationally efficient.}
\label{tab:dynamic_regret_comparison}
\vspace{8pt}
\small
\setlength{\tabcolsep}{2pt}
\renewcommand{\arraystretch}{1.35}
\begin{tabular*}{\linewidth}{@{\extracolsep{\fill}}
  >{\centering\arraybackslash}m{0.200\linewidth}|
  >{\centering\arraybackslash}m{0.210\linewidth}
  >{\centering\arraybackslash}m{0.210\linewidth}
  >{\centering\arraybackslash}m{0.210\linewidth}|
  >{\centering\arraybackslash}m{\dimexpr0.170\linewidth-9\tabcolsep-2\arrayrulewidth\relax}}
\toprule
\multirow[c]{2}{=}{\centering\textbf{Method}}
  & \multicolumn{3}{c|}{\textbf{Regret Bounds}}
  & \multirow[c]{2}{=}{\centering\textbf{Efficient}} \\
\noalign{\vskip-\aboverulesep}
\cmidrule[\arrayrulewidth](l{3pt}r{3pt}){2-4}
\noalign{\vskip-\dimexpr\belowrulesep+\arrayrulewidth\relax}
  & cvx & str-cvx & exp-concave & \\
\hline
\citet{NeurIPS:2018:Zhang}
  & $\gO(\sqrt{T(1+P_T)})$
  & --
  & --
  & \rightsym \\
\citet{COLT:2021:Baby}
  & --
  & $\widetilde{\gO}(d^{7/3}T^{1/3}P_T^{2/3})$
  & $\widetilde{\gO}(d^{23/6}T^{1/3}P_T^{2/3})$
  & \rightsym \\
\citet{AISTATS:2022:Baby}
  & --
  & $\widetilde{\gO}(d^{2/3}T^{1/3}P_T^{2/3})$
  & $\widetilde{\gO}(d^{10/3}T^{1/3}P_T^{2/3})$
  & \rightsym  \\
\citet{ICML:2025:Zhang}
  & --
  & --
  & $\widetilde{\gO}(dT^{1/3}P_T^{2/3})$
  & \wrongsym \\
\hline
\textbf{This work}
  & $\gO(\sqrt{T(1+P_T)})$
  & $\widetilde{\gO}(T^{1/3}P_T^{2/3})$
  & $\widetilde{\gO}(d^{2/3}T^{1/3}P_T^{2/3})$
  & \rightsym \\
\bottomrule
\end{tabular*}
\vspace{-10pt}
\end{table}

In this section, we briefly review both dynamic and switching regret minimization in OCO.

\noindent
\textbf{Dynamic regret minimization.}
In the literature, there are two different forms of dynamic regret. One is the general case \eqref{eq:dynamic_regret} proposed by \citet{ICML:2003:Zinkevich}, who introduces the path length
\begin{equation}
    \label{eq:pt}
    P_T = \sum\nolimits_{t=2}^T\|\vu_t-\vu_{t-1}\|_2,
\end{equation}
to establish the first dynamic regret bound of $\gO(\sqrt{T}(1+P_T))$ for Online Gradient Descent (OGD). Subsequently, \citet{NeurIPS:2018:Zhang} improve this result to the minimax-optimal $\gO(\sqrt{T(1+P_T)})$ bound by aggregating multiple OGD instances via the Hedge Algorithm \citep{Others:1997:Freund}. To exploit the curvature of losses, \citet{COLT:2021:Baby} develop analyses based on the KKT conditions of offline optimal comparator sequence, and establish $\widetilde{\gO}(d^{7/3}T^{1/3}P_T^{2/3})$ and $\widetilde{\gO}(d^{23/6}T^{1/3}P_T^{2/3})$ bounds for strongly convex losses and exp-concave losses, respectively. Later, following similar analyses, \cite{AISTATS:2022:Baby} achieve improved bounds of $\widetilde{\gO}(d^{2/3}T^{1/3}P_T^{2/3})$ for strongly convex losses and $\widetilde{\gO}(d^{10/3}T^{1/3}P_T^{2/3})$ for exp-concave losses, with tighter dimension dependences. Recently, \citet{ICML:2025:Zhang} develop a mixability-based analysis for dynamic regret minimization and extend it to exp-concave losses, obtaining an $\widetilde{\gO}(dT^{1/3}P_T^{2/3})$ bound with linear dimension dependence. However, their method requires computationally expensive KL projections onto constrained Gaussian mixtures. We present comparisons of these methods in Table~\ref{tab:dynamic_regret_comparison}.

Another line of research focuses on the \emph{worst-case} dynamic regret, in which the comparators in \eqref{eq:dynamic_regret} are chosen as the per-round minimizers $\vu_t^\star\in\argmin_{\vu\in\X}f_t(\vu)$. To quantify the difficulty of tracking these minimizers, existing studies establish regret bounds in terms of the temporal variation of the loss functions or the minimizer sequence \citep{Others:2015:Besbes,AISTATS:2015:Jadbabaie,ICML:2016:Yang,CDC:2016:Mokhtari,NeurIPS:2017:Zhang,NeurIPS:2019:Baby,AAAI:2021:Wan:B,COLT:2023:Wan,NeurIPS:2024:Xu}. In this paper, we focus on the general form of dynamic regret \eqref{eq:dynamic_regret}, and aim to establish the dynamic regret bounds that adapt to the path-length $P_T$ of \textit{any} comparator sequence.

\noindent
\textbf{Switching regret minimization.}
Switching regret is defined as the difference between the cumulative loss incurred by the learner and that of the best piecewise-constant comparator sequence \citep{NeurIPS:2024:Pasteris}. Formally, let $\gS=\{\gI_1,\ldots,\gI_{|\gS|}\}$ be an arbitrary partition of the time horizon $T$ into contiguous intervals. The switching regret with respect to $\gS$ is defined as
\begin{equation}
\label{eq:switching_regret_partition}
  \SW_T(\gS)
  =\sum\nolimits_{\gI \in \gS}\left[\sum\nolimits_{t \in \gI}f_t(\vx_t)
   -\min_{\vx\in\X}\sum\nolimits_{t\in\gI}f_t(\vx)\right],
\end{equation}
which is the sum of the static regret over each interval $\gI\in\gS$. The comparator remains fixed within each interval and switches at most $|\gS|-1$ times. The learner does not know the segmentation in advance, and its switching regret guarantee should hold simultaneously for all possible segmentations.

To minimize \eqref{eq:switching_regret_partition}, \citet{NeurIPS:2024:Pasteris} propose RESET for general convex losses, which recursively aggregates OGD experts over a segment tree and achieves an $\gO(\sum_{\gI\in\gS}\sqrt{|\gI|})$ bound. Later, \citet{ICML:2026:Yang} propose IRESET, which exploits loss curvature by combining a meta-algorithm with a second-order bound and suitable expert algorithms. With OGD as the expert algorithm for strongly convex losses and Online Newton Step (ONS) \citep{Others:2007:Hazan} for exp-concave losses, IRESET attains switching-regret bounds of $\gO(\sum_{\gI\in\gS}\log^2(|\gI|))$ and $\gO(d\sum_{\gI\in\gS}\log^2(|\gI|))$, respectively.
\section{Reduction via an Unbiased Random Sequence}
\label{sec:reduction}

In this section, we first  introduce necessary preliminaries, including  assumptions and definitions. Then, we present the reduction from  dynamic regret minimization to switching regret minimization.

\subsection{Preliminaries}
\label{sec:preliminaries}

Similar to previous studies in OCO, we list  standard assumptions and basic definitions below.

\begin{assumption}
    \label{assump:K-bound}
    The convex set $\X$ contains $\mathbf{0}$ and belongs to a ball  with radius $R$, i.e., $\X \subseteq B(\mathbf{0}, R)$.
\end{assumption}

\begin{assumption}
    \label{assump:Lipschitz}
    At each round $t$, the loss function $f_{t}(\cdot)$ is $G$-Lipschitz over $\X$, i.e., 
    \begin{equation}
        \forall \vx,\vy \in \X,~|f_{t}(\vx) - f_{t}(\vy)| \leq G \| \vx - \vy\|_2.
        \nonumber
    \end{equation}
\end{assumption}

\begin{definition}
    \label{def:convex}
    A function $f:\X\to\R$ is convex if
    \begin{equation*}
        \forall \vx, \vy \in \X,~ f(\vy) \geq f(\vx) + \nabla f(\vx)^\top (\vy - \vx).
    \end{equation*}
\end{definition}

\begin{definition}
    \label{def:strongly-convex}
    A  function $f:\X\to\R$ is $\lambda$-strongly convex if
    \begin{equation*}
        \forall \vx, \vy \in \X,~ f(\vy) \geq f(\vx) + \nabla f(\vx)^\top (\vy - \vx) + \frac{\lambda}{2} \|\vx - \vy\|_2^2.
    \end{equation*}
\end{definition}

\begin{definition}
\label{def:exp_concavity}
A function $f:\X\to\R$ is $\alpha$-exp-concave if $\exp(-\alpha f(\cdot))$ is concave on $\X$.
\end{definition}

\subsection{Reducing Dynamic Regret to Switching Regret}
\label{sec:motivation}

To clarify our motivation, we first examine the reduction from dynamic regret to switching regret based on a deterministic sequence, but obtain suboptimal guarantees. Motivated by its limitation, we then present our reduction that ensures optimal bounds for three types of losses.

\noindent
\textbf{Deterministic sequence.} From
\eqref{eq:dynamic_regret} and \eqref{eq:switching_regret_partition}, we observe that switching regret is a special case of dynamic regret, where  comparators are restricted to a sequence with few changes. Hence, a natural idea is to reduce dynamic regret to switching regret by replacing the original comparator sequence with a deterministic \textit{piecewise-constant} one. On the one hand, the piecewise-constant sequence should have few switches to ensure a small switching regret bound. On the other hand, it should stay close to the original sequence to keep the approximation error manageable. However, balancing these goals yields suboptimal bounds, as illustrated below for strongly convex losses.

Let $\vz_{1:T}$ be a piecewise-constant approximation of $\vu_{1:T}$ with tolerance $\varepsilon>0$, and let $J(\vz_{1:T})$ denote its number of switches. We initialize $\vz_1=\vu_1$. For each $t\ge2$, we set $\vz_t=\vz_{t-1}$ if $\norm{\vz_{t-1}-\vu_t}_2\le\varepsilon$; otherwise, we set $\vz_t=\vu_t$. This construction ensures $J(\vz_{1:T})\le P_T/\varepsilon$ and $\norm{\vz_t-\vu_t}_2\le\varepsilon$ at every round. By adding and subtracting $f_t(\vz_t)$ at each round, we obtain
\begin{equation}
  \label{eq:dynamic_decompose_z}
  \begin{aligned}
    \DReg_T(\vu_{1:T})
    &= \underbrace{
      \sum\nolimits_{t=1}^T \bigl[f_t(\vx_t)-f_t(\vz_t)\bigr]
    }_{\text{switching regret}}
    + \underbrace{
      \sum\nolimits_{t=1}^T \bigl[f_t(\vz_t)-f_t(\vu_t)\bigr]
    }_{\text{approximation error}},
  \end{aligned}
\end{equation}
where the first term represents the \emph{switching regret} against the comparator sequence $\vz_{1:T}$, and the second term is the \emph{approximation error} incurred by approximating $\vu_{1:T}$ with $\vz_{1:T}$. To bound the first term, we run an existing switching-regret minimization algorithm for strongly convex losses on $f_1,\ldots,f_T$, obtaining the following bound \citep{ICML:2026:Yang}:
\begin{equation}
     \label{eq:str_cvx_switching}
     \sum\nolimits_{t=1}^T \bigl[f_t(\vx_t)-f_t(\vz_t)\bigr] = \gO \bigl( (1+J(\vz_{1:T}) ) \log^2 T \bigr) =  \gO  \bigl((1 + P_T / \varepsilon)\log^2 T \bigr),
\end{equation}
which depends on the number of switches $J(\vz_{1:T})$ and becomes tighter as this number decreases. To bound the second term, we use the Lipschitz continuity of $f_t$ in Assumption~\ref{assump:Lipschitz} to obtain
\begin{equation}
  \label{eq:ap_z}
  \sum\nolimits_{t=1}^T \bigl[f_t(\vz_t)-f_t(\vu_t)\bigr] \leq  \sum\nolimits_{t=1}^T G \norm{\vz_t - \vu_t}_2 \leq GT \varepsilon.
\end{equation}

Substituting \eqref{eq:str_cvx_switching} and \eqref{eq:ap_z} into \eqref{eq:dynamic_decompose_z}, we obtain
\begin{equation}
  \label{eq:str_cvx_z}
  \DReg_T(\vu_{1:T}) =  \gO \left(\left(1 + \frac{P_T}{\varepsilon}\right) \log^2 T + GT \varepsilon\right) = \widetilde{\gO}(\sqrt{T P_T}),
\end{equation}
where we set the tolerance $\varepsilon = \widetilde{{\gO}}( \sqrt{P_T/T})$. This result is worse than the optimal $\widetilde{\gO}(T^{1/3}P_T^{2/3})$ bound for strongly convex losses \citep{COLT:2021:Baby}.  The same analysis also yields suboptimal bounds of $\gO (T^{2/3}P_T^{1/3})$ and $\widetilde{\gO}(\sqrt{d T P_T})$ for general convex and exp-concave losses, respectively. The essential limitation is that the deterministic construction does not yield a favorable trade-off between the switching regret and the approximation error.

\textbf{Technical contributions.}
To overcome this limitation, we exploit the curvature of $f_t$ to construct surrogate losses whose regret upper-bounds the original regret. The resulting surrogate losses admit a quadratic expansion around the original comparator, allowing us to decompose the approximation error into a \emph{linear} term and a \emph{quadratic} term. We then introduce an auxiliary random sequence that is unbiased at each round, with its variance bounded by $\varepsilon^2$ and the expected number of switches bounded by $P_T/\varepsilon$. The unbiasedness eliminates the linear term in expectation, and the variance controls the quadratic term, yielding an $\gO(T\varepsilon^2)$ approximation error. By choosing $\varepsilon$ appropriately to balance this error with the switching-regret bound, we obtain the optimal dynamic regret bound up to logarithmic factors. Below, we present these two ingredients and the resulting reduction.

We define the surrogate losses $\ell_t$ on an enlarged domain $\Y\supseteq\X$ and require them to be $H$-smooth:
\begin{equation}
  \label{eq:surrogate_loss}
  \ell_t(\vv)
  \le \ell_t(\vu)
    +\inner{\nabla\ell_t(\vu)}{\vv-\vu}
    +\frac{H}{2}\norm{\vv-\vu}_2^2
\end{equation}
for all $\vu,\vv\in\Y$ and every $t$, with a common constant $H\ge0$. The following lemma establishes the existence of an auxiliary random sequence with the desired properties.
\begin{lemma}
    \label{lem:unbiased_sequence}
    Let $\X\subseteq B(\mathbf{0},R)$ and $\Y=B(\mathbf{0},2R)$.
    For any comparator sequence $\vu_{1:T}\in\X$ and any
    $0<\varepsilon\le R$, we can construct an auxiliary random
    sequence $\vV_{1:T}$ taking values in $\Y$ such that
    \begin{equation*}
        \E[\vV_t]=\vu_t,\qquad
        \E\left[\norm{\vV_t-\vu_t}_2^2\right]\le\varepsilon^2,\qquad
        \E[J(\vV_{1:T})]\le\frac{P_T}{\varepsilon},
    \end{equation*}
    where the first two properties hold for every round and the expectations are taken over  $\vV_{1:T}$.
\end{lemma}
Note that the random sequence in Lemma~\ref{lem:unbiased_sequence} may lie outside the original domain $\X$, so we define the surrogate losses on the enlarged domain $\Y$, which provably controls of the original dynamic regret. Specifically, by projecting predictions of the learner onto $\X$ and applying a suitable gradient correction, the dynamic regret under $f_t$ is upper-bounded by that under $\ell_t$ against the same comparator sequence \citep{COLT:2018:Cutkosky}.  We provide the detailed construction in Section~\ref{sec:method}. In the following, we hence focus on analyzing the dynamic regret of surrogate losses over $\Y$.

Let $\vy_t\in\Y$ denote the predictions of a switching-regret minimization algorithm run on the surrogate losses $\ell_1,\ldots,\ell_T$. Analogously to \eqref{eq:dynamic_decompose_z}, we use the auxiliary random sequence $\vV_{1:T}$ from Lemma~\ref{lem:unbiased_sequence} as an intermediate comparator. By adding and subtracting $\ell_t(\vV_t)$ at each round and taking expectations over the randomness of $\vV_{1:T}$, we decompose the dynamic regret under the surrogate losses into the expected switching regret against $\vV_{1:T}$ and the expected approximation error:
\begin{equation}  \label{eq:dy:sl}
    \sum\nolimits_{t=1}^T \bigl[\ell_t(\vy_t)-\ell_t(\vu_t)\bigr] = \E\left[\sum\nolimits_{t=1}^T \bigl[\ell_t(\vy_t)-\ell_t(\vV_t)\bigr]\right] +\E\left[\sum\nolimits_{t=1}^T \bigl[\ell_t(\vV_t)-\ell_t(\vu_t)\bigr]\right].
\end{equation}
For $\lambda$-strongly convex losses, the corresponding surrogate losses are also $\lambda$-strongly convex. We can therefore run the existing switching-regret minimization algorithm \citep{ICML:2026:Yang} on the surrogate loss sequence $\ell_1,\ldots,\ell_T$ to obtain a bound analogous to \eqref{eq:str_cvx_switching}:
\begin{equation}
  \label{eq:sw}
 \E\left[\sum\nolimits_{t=1}^T \bigl[\ell_t(\vy_t)-\ell_t(\vV_t)\bigr]\right] \le \E\left[ \gO  \bigl( (1+J(\vV_{1:T}) ) \log^2 T \bigr)  \right] =  \gO \bigl((1 + P_T / \varepsilon)\log^2 T \bigr).
\end{equation}
By the property of $\ell_t$ in \eqref{eq:surrogate_loss}, the approximation error is bounded by
\begin{equation}
  \label{eq:ap}
  \E\left[\sum_{t=1}^T
    \bigl[\ell_t(\vV_t)-\ell_t(\vu_t)\bigr]\right] \le \E\left[\sum_{t=1}^T
    \inner{\nabla\ell_t(\vu_t)}{\vV_t-\vu_t}\right]
    +\frac{H}{2}\E\left[\sum_{t=1}^T
    \norm{\vV_t-\vu_t}_2^2\right] \leq \frac{HT\varepsilon^2}{2},
\end{equation}
where the last inequality follows from Lemma~\ref{lem:unbiased_sequence}: the linear term vanishes by unbiasedness, i.e., $\E[\vV_t-\vu_t]=\mathbf{0}$, and the quadratic term is controlled by the variance bound, i.e., $\E[\norm{\vV_t-\vu_t}_2^2]\le\varepsilon^2$.

Combining \eqref{eq:dy:sl}, \eqref{eq:sw} and \eqref{eq:ap} delivers
\begin{equation}
 \sum_{t=1}^T \bigl[\ell_t(\vy_t)-\ell_t(\vu_t)\bigr]  = \gO \left( \left(1+\frac{P_T}{\varepsilon}\right)\log^2 T +H T\varepsilon^2  \right) = \widetilde{\gO}\left(T^{1/3}P_T^{2/3}\right),
\end{equation}
where we set the tolerance $\varepsilon = \widetilde{{\gO}}(T^{-1/3}P_T^{1/3})$. Note that this bound attains the optimal guarantee for strongly convex losses, up to logarithmic factors  \citep{COLT:2021:Baby}.

The following theorem formalizes this reduction for any learner with a switching-regret guarantee.
\begin{theorem}
    \label{thm:reduction}
    Let $\X\subseteq B(\mathbf{0},R)$ and $\Y=B(\mathbf{0},2R)$ for some $R>0$. Suppose the surrogate losses $\ell_t:\Y\to\R$ satisfy \eqref{eq:surrogate_loss} with some $H\ge0$. An online learner run on $\ell_1,\ldots,\ell_T$ with predictions $\vy_t\in\Y$  guarantees
    \begin{equation*}
        \sum\nolimits_{t=1}^T
        \bigl[\ell_t(\vy_t)-\ell_t(\vv_t)\bigr]
        \le \Phi\bigl(T,J(\vv_{1:T})\bigr)
    \end{equation*}
    for any comparator sequence $\vv_{1:T}\in\Y$, where $J(\vv_{1:T})$ denotes the number of switches in $\vv_{1:T}$, and $\Phi(T,J)$ is the switching regret bound of the learner over $T$ rounds against comparator sequences with $J$ switches. Then, for any comparator sequence $\vu_{1:T}\in\X$ and any $0<\varepsilon\le R$, we have
    \begin{equation}
        \label{eq:reduction_bound}
        \sum_{t=1}^T \bigl[\ell_t(\vy_t)-\ell_t(\vu_t)\bigr] \le \E\bigl[\Phi(T,J(\vV_{1:T}))\bigr] +\frac{HT\varepsilon^2}{2},
    \end{equation}
    where $\vV_{1:T}$ is the auxiliary random sequence constructed in Lemma~\ref{lem:unbiased_sequence}. The expectation is taken only over its auxiliary randomness, with surrogate losses and predictions of the learner held fixed.
\end{theorem}

\noindent
\textbf{Remark.}
Theorem~\ref{thm:reduction} provides a unified reduction for all three loss classes. For $\lambda$-strongly convex and $\alpha$-exp-concave losses, the surrogate losses in Sections~\ref{sec:strongly_convex} and~\ref{sec:exp_concave} satisfy \eqref{eq:surrogate_loss} with $H=\lambda$ and $H=\beta G^2$, respectively, yielding an $\gO(T\varepsilon^2)$ approximation error. For general convex losses, the surrogate loss in Section~\ref{sec:convex} satisfies the same condition with $H=0$, so the approximation error vanishes. In each case, the dynamic regret bound follows by substituting the corresponding switching-regret guarantee into \eqref{eq:reduction_bound} and choosing $\varepsilon$ appropriately. The instantiations below yield bounds that are optimal in $T$ and $P_T$, up to logarithmic factors, through this common reduction.

\section{Methods}
\label{sec:method}

We now instantiate the reduction in Theorem~\ref{thm:reduction} for strongly convex, exp-concave, and general convex losses. To implement the reduction, we construct surrogate losses on the enlarged domain $\Y$ using gradient feedback at feasible decisions in $\X$. These losses retain the curvature needed by the switching-regret learner, satisfy \eqref{eq:surrogate_loss}, and ensure that their regret upper-bounds the original regret. This allows us to translate the guarantees from Section~\ref{sec:motivation} into dynamic regret bounds for the actual decisions. The auxiliary random sequence is used only in the analysis and need not be generated.

Algorithm~\ref{alg:proper_learning} summarizes the procedure. We first select the enlarged domain $\Y$ and the switching-regret learner $\mathcal{L}$ according to the loss type, and initialize $\mathcal{L}$ on $\Y$ (Lines~\ref{line:s2d-configure}--\ref{line:s2d-initialize}).\footnote{Since strongly adaptive regret bounds can imply switching regret guarantees \citep{NeurIPS:2024:Pasteris}, one could also instantiate the learner $\mathcal{L}$ with an algorithm that minimizes strongly adaptive regret. However, this strategy may lead to suboptimal dynamic regret bounds. In particular, for general convex losses, an $\gO(\sqrt{|\gI|}\log T)$ strongly adaptive regret bound on any interval $\gI$ yields an $\gO(\sqrt{T(1+P_T)}\log T)$ dynamic regret bound via our framework, which is worse than \eqref{eq:convex_dynamic} in  Theorem~\ref{thm:cov_dynamic} by an additional $\log T$ factor.} At each round $t$, we obtain an auxiliary prediction $\vy_t\in\Y$ from $\mathcal{L}$ (Line~\ref{line:s2d-predict}), play its projection $\vx_t=\Pi_{\X}(\vy_t)$, and observe $\nabla f_t(\vx_t)$ (Line~\ref{line:s2d-play}). We then construct the surrogate loss $\ell_t$ for the given loss type and update $\mathcal{L}$ with it, using normalization when required (Lines~\ref{line:s2d-surrogate}--\ref{line:s2d-update}). Below, we specify the configurations and present the corresponding dynamic regret guarantees for the three loss classes.

\subsection{Strongly Convex Losses}
\label{sec:strongly_convex}

\begin{algorithm}[t]
\caption{Switching-to-Dynamic Regret Minimization}
\label{alg:proper_learning}
\algrenewcommand{\algorithmicrequire}{\textbf{Input:}}
\begin{algorithmic}[1]
\Require Horizon $T$, domain $\X$, bounds $R$ and $G$,
loss type, and curvature parameter (if applicable).

\State Select the enlarged domain $\Y$ and the
switching-regret learner $\mathcal{L}$ according to the loss type.
\phantomsection\label{line:s2d-configure}

\State Initialize $\mathcal{L}$ on $\Y$.
\phantomsection\label{line:s2d-initialize}

\For{$t=1,\ldots,T$} \phantomsection\label{line:s2d-loop}
  \State Obtain the auxiliary prediction $\vy_t\in\Y$ from $\mathcal{L}$. \phantomsection\label{line:s2d-predict}
  \State Play $\vx_t=\Pi_{\X}(\vy_t)$ and observe $\nabla f_t(\vx_t)$. \phantomsection\label{line:s2d-play}
\State Construct the surrogate loss $\ell_t$ as specified in
Sections~\ref{sec:strongly_convex}--\ref{sec:convex}.
\phantomsection\label{line:s2d-surrogate}
  \State Update $\mathcal{L}$ with $\ell_t$. \phantomsection\label{line:s2d-update}
\EndFor \phantomsection\label{line:s2d-end}
\end{algorithmic}
\end{algorithm}

Initially, we consider the case where  $f_t$ is $\lambda$-strongly convex, and  choose the surrogate loss 
\begin{equation}
    \label{eq:surrogate_str_cvx}
    \ell_t^{sc}(\vy)=\inner{\vh_t}{\vy-\vy_t} +\frac{\lambda}{2}\norm{\vy-\vx_t}_2^2,
\end{equation}
where the corrected gradient is set as \citep{ICML:2020:Cutkosky}
\[
  \vh_t=\nabla f_t(\vx_t)-\ind\{c_t<0\}\,c_t\vn_t,
\]
with $c_t=\inner{\nabla f_t(\vx_t)}{\vn_t}$ and 
\begin{equation}
  \vn_t=\begin{cases}
    (\vy_t-\vx_t)/\norm{\vy_t-\vx_t}_2,&\vy_t\ne\vx_t,\\
    0,&\vy_t=\vx_t.
  \end{cases}
  \nonumber
\end{equation}
It can be readily verified that \eqref{eq:surrogate_str_cvx} satisfies \eqref{eq:surrogate_loss} with $H=\lambda$ and enjoys the following property:
\begin{equation}
  \label{eq:f_l}
  f_t(\vx_t)-f_t(\vu) \le\ell_t(\vy_t)-\ell_t(\vu)
\end{equation}
for every $\vu\in\X$ and $\vy_t\in\Y$, with $\vx_t=\Pi_{\X}(\vy_t)$ \citep{NeurIPS:2024:Yang}. Therefore, we can conveniently derive dynamic regret bounds for the original losses $f_t$ by analyzing that for the surrogate losses $\ell_t$.

Next, we examine the diameter of the domains $\X$ and $\Y$. For any $\vx,\vu\in\X$, the $\lambda$-strong convexity of $f_t$ and Assumption~\ref{assump:Lipschitz} imply that
\begin{equation*}
\begin{aligned}
  \lambda\|\vx-\vu\|_2^2
  &\le
  \langle \nabla f_t(\vx)-\nabla f_t(\vu),\vx-\vu\rangle \le 2G\|\vx-\vu\|_2,
\end{aligned}
\end{equation*}
which indicates $\diam(\X)\le 2G/\lambda$. In other words, $\X$ is contained in a ball $B(\mathbf{0}, 2G/\lambda)$ and we choose the enlarged set $\Y = B(\mathbf{0}, 4G/\lambda)$ with the radius $4G/\lambda$. We then bound the gradient norm of the surrogate loss in \eqref{eq:surrogate_str_cvx}. Since $\norm{\vh_t}_2\le G$ and $\vx_t\in\X\subseteq\Y$, for any $\vy\in\Y$ we have
\[
  \norm{\nabla\ell_t^{sc}(\vy)}_2
  =\norm{\vh_t+\lambda(\vy-\vx_t)}_2
  \le G+\lambda \cdot 8G / \lambda=9G.
\]
With these configurations, we instantiate the switching-regret learner $\mathcal{L}$ with IRESET-OGD \citep{ICML:2026:Yang}, and present the dynamic regret bound for strongly convex losses below.

\begin{theorem}
    \label{thm:sc_dynamic}Under Assumptions~\ref{assump:K-bound}~and~\ref{assump:Lipschitz}, suppose $f_t$ is $\lambda$-strongly convex. With the configurations above, Algorithm~\ref{alg:proper_learning} ensures, for every comparator sequence $\vu_{1:T}\in\X$,
    \begin{equation}
    \DReg_T(\vu_{1:T}) = \gO \left(
        \frac{G^2}{\lambda}(1+\log T)
        +\left(\frac{G^4TP_T^2}{\lambda}\right)^{1/3}
        (1+\log T)^{4/3}  \right)= \widetilde{\gO}\left(T^{1/3}P_T^{2/3}\right).
    \nonumber
    \end{equation}
\end{theorem}
\noindent
\textbf{Remark.}
Theorem~\ref{thm:sc_dynamic} matches the minimax optimal bound, up to logarithmic factors \citep{COLT:2021:Baby} and eliminates the dimension dependence in previous results \citep{AISTATS:2022:Baby}. Notably, our analysis uses the existing switching-regret guarantee of IRESET-OGD as a black box and combines it with the properties of the auxiliary random sequence in Lemma~\ref{lem:unbiased_sequence}. This yields a simple proof without analyzing the structure of an offline optimal comparator sequence.

\subsection{Exp-Concave Losses}
\label{sec:exp_concave}

In this part, we consider the case where  $f_t$ is $\alpha$-exp-concave, and choose the  surrogate loss:
\begin{equation}
\label{eq:exp_surrogate}
  \ell_t^{\mathrm{exp}}(\vy)
  = \inner{\vh_t}{\vy-\vy_t}
    + \frac{\beta}{2}\inner{\vh_t}{\vy-\vy_t}^2,
\end{equation}
in which $\beta= \min\{1/(32GR),\alpha/2\}$. We can also verify that \eqref{eq:exp_surrogate} satisfies \eqref{eq:surrogate_loss} with $H=\beta G^2$ and \eqref{eq:f_l} \citep{ICML:2019:Mhammedi}. Moreover, the value of $\beta$ delivers $R\le 1/(32\beta G)$ and hence $\X\subseteq B(\mathbf{0},1/(32\beta G))$. We therefore set the enlarged domain to $\Y=B(\mathbf{0},1/(16\beta G))$, with the radius $1/(16\beta G)$. Since $\norm{\vh_t}_2\le G$, we have
\begin{equation*}
  \begin{aligned}
    \norm{\nabla\ell_t^{\mathrm{exp}}(\vy)}_2
    &=\left|1+\beta\inner{\vh_t}{\vy-\vy_t}\right|\norm{\vh_t}_2 
    \le \left(1+\beta G \cdot \frac{1}{8\beta G} \right)G =\frac{9}{8}G\le\sqrt{2}G.
  \end{aligned}
\end{equation*}
Instantiating the switching-regret learner with IRESET-ONS \citep{ICML:2026:Yang} yields the following dynamic regret bound for exp-concave losses.

\begin{theorem}
\label{thm:exp_dynamic}
Under Assumptions~\ref{assump:K-bound}~and~\ref{assump:Lipschitz}, suppose $f_t$ is $\alpha$-exp-concave. With the configurations above, Algorithm~\ref{alg:proper_learning} ensures, for every comparator sequence $\vu_{1:T}\in\X$,
\begin{equation}
  \begin{aligned}
    \DReg_T(\vu_{1:T})
    &=\gO\!\left(\frac{d}{\beta}(1+\log T)
      +\left(\frac{d^2G^2TP_T^2}{\beta}\right)^{1/3}(1+\log T)^{4/3}\right) =\widetilde{\gO}\!\left(d^{2/3}T^{1/3}P_T^{2/3}\right).
  \end{aligned}
  \nonumber
\end{equation}
\end{theorem}

\noindent
\textbf{Remark.}
Theorem~\ref{thm:exp_dynamic} achieves the minimax-optimal bound for exp-concave losses, up to additional logarithmic factors \citep{COLT:2021:Baby}. Notably, the dependence on $d$ is inherited from the switching regret guarantee of IRESET with ONS experts, and our reduction introduces no additional dependence on $d$. In other words, any improvement in the dimension dependence of the switching regret guarantee can be carried over to the dynamic regret bound through our reduction.

\noindent
\textbf{Remark.}
Compared with \citet{ICML:2025:Zhang}, Theorem~\ref{thm:exp_dynamic} improves the dimension dependence from $d$ to $d^{2/3}$. Moreover, our framework requires only running a single instance of \textit{any} existing switching-regret algorithm  over the surrogate loss \eqref{eq:exp_surrogate}, with its predictions projected onto $\X$. This construction avoids the complicate KL projections and makes it easier to understand and implement.

\subsection{General Convex Losses}
\label{sec:convex}

We further investigate the case where   $f_t$ is general convex, and use the following surrogate loss: 
\begin{equation}
  \label{eq:convex_surrogate}
  \ell_t^{cov}(\vy)=\inner{\vh_t}{\vy-\vy_t},
\end{equation}
which also satisfies \eqref{eq:surrogate_loss} with $H=0$, as well as \eqref{eq:f_l} \citep{ICML:2020:Cutkosky}. Moreover, its gradient satisfies $\norm{\nabla\ell_t^{\mathrm{cov}}(\vy)}_2 =\norm{\vh_t}_2\le G$ by Assumption~\ref{assump:Lipschitz}. Therefore, we set the enlarged domain $\Y=B(\mathbf{0},2R)$ and normalize the surrogate losses to $[0,1]$. Then, instantiating the switching-regret learner with RESET \citep{NeurIPS:2024:Pasteris} yields the following dynamic regret bound.

\begin{theorem}
\label{thm:cov_dynamic}
Under Assumptions~\ref{assump:K-bound} and~\ref{assump:Lipschitz}, suppose $f_t$ is general convex. With the configuration above, Algorithm~\ref{alg:proper_learning} ensures, for every comparator sequence $\vu_{1:T}\in\X$,
\begin{equation}
  \label{eq:convex_dynamic}
    \DReg_T(\vu_{1:T}) = \gO\left(G\sqrt{T(R^2+RP_T)}\right) = \gO\left(\sqrt{T(1+P_T)}\right).
  \end{equation}
\end{theorem}

\noindent
\textbf{Remark.}
Theorem~\ref{thm:cov_dynamic} matches the optimal result for general convex losses \citep{NeurIPS:2018:Zhang}.  Theorems~\ref{thm:sc_dynamic}, \ref{thm:exp_dynamic}, and~\ref{thm:cov_dynamic} together show that the same reduction applies to all three types of losses with optimal dynamic regret bounds, highlighting the simplicity and generality of our framework.

\section{Theoretical Analysis}
\label{sec:theoretical_analysis}

Due to the limitation of space, we only provide the proof of Lemma~\ref{lem:unbiased_sequence}and Theorem~\ref{thm:sc_dynamic}. The omitted proofs can be found in the supplementary material. Throughout these proofs, all expectations are taken over the auxiliary random sequence $\vV_{1:T}$ constructed in Lemma~\ref{lem:unbiased_sequence}.

\subsection{Proof of Lemma~\ref{lem:unbiased_sequence}}
\label{app:proof_lemma_random}

We first show how to construct an auxiliary random sequence $\vV_{1:T}$ for the comparator sequence $\vu_{1:T}$. The basic idea is to keep the auxiliary sequence unchanged with high probability when $\vu_{1:T}$ moves only slightly. When an update occurs, we move to a point that compensates for this small update probability. Specifically, given $\varepsilon>0$, we initialize $\vV_1=\vu_1$ and define, for each $t\ge2$, 
\[
  \vdelta_t=\vu_t-\vu_{t-1},\qquad
  \Delta_t=\|\vdelta_t\|_2,\qquad
  p_t=\min\{1,\Delta_t/\varepsilon\}.
\]
If $\Delta_t=0$, we keep $\vV_t=\vV_{t-1}$. Otherwise, independently of previous random choices, we set
\begin{equation}
  \label{eq:reset_construction}
  \vV_t=\begin{cases}
    \vu_{t-1}+\vdelta_t/p_t,&\text{with probability }p_t,\\
    \vV_{t-1},&\text{with probability }1-p_t.
  \end{cases}
\end{equation}
When $0<\Delta_t<\varepsilon$, the sequence updates with probability $p_t = \Delta_t/\varepsilon$ to a point at distance $\varepsilon$ from $\vu_{t-1}$ in the direction of $\vdelta_t$. The factor $1/p_t$ in \eqref{eq:reset_construction} makes up for the small update probability, so that the mean still moves by $\vdelta_t$. When $\Delta_t\ge\varepsilon$, we simply set $\vV_t=\vu_t$.

We now state the properties of this construction. Let $J(\vV_{1:T})=\sum_{t=2}^T\ind\{\vV_t\ne\vV_{t-1}\}$ denote the number of switches, so that the sequence $\vV_{1:T}$ has $J(\vV_{1:T})+1$ constant pieces. Let $B(\mathbf{0},R)=\{\vx\in\R^d:\norm{\vx}_2\le R\}$ denote the Euclidean ball of radius $R$. Since the update point can lie beyond $\vu_t$, we allow the random sequence to take values in a larger ball than the original comparator.

Next, we show that every  $\vV_t$ lies in the enlarged domain $\Y = B(\mathbf{0},2R)$. The initial point $\vV_1=\vu_1$ lies in $\X \subseteq B(\mathbf{0},R)$. At an update, the new point is either $\vu_t$ when $p_t=1$, or a point at distance $\varepsilon\le R$ from $\vu_{t-1}$ when $0<p_t<1$. In both update cases $\vV_t$ lies in $B(\mathbf{0},2R)$. If no update occurs at round $t$, this property still holds, since $\vV_t = \vV_{t-1} \in B(\mathbf{0}, 2R)$.

Then, we prove the unbiasedness by induction. The claim holds at $t=1$. If $p_t=0$, both the comparator and the random sequence remain unchanged. If $p_t>0$, the independence of the new random choice and the induction hypothesis give
\[
  \E[\vV_t]
  =(1-p_t)\vu_{t-1}
    +p_t\left(\vu_{t-1}+\frac{\vdelta_t}{p_t}\right)
  =\vu_t.
\]

To bound the variance, let $W_t=\E[\norm{\vV_t-\vu_t}_2^2]$, with $W_1=0$. If $p_t=0$, then $W_t=W_{t-1}$. For $p_t>0$, expanding the two cases in \eqref{eq:reset_construction} and using $\E[\vV_{t-1}-\vu_{t-1}]=\mathbf{0}$ yield
\begin{equation*}
  \begin{aligned}
    W_t
    &=(1-p_t)(W_{t-1}+\Delta_t^2)
      +p_t\left(\frac{1-p_t}{p_t}\right)^2\Delta_t^2 
    =(1-p_t)W_{t-1}+\frac{1-p_t}{p_t}\Delta_t^2.
  \end{aligned}
\end{equation*}
If $p_t=1$, then $W_t=0$. If $0<p_t<1$, then $\Delta_t=p_t\varepsilon$. Thus, assuming $W_{t-1}\le\varepsilon^2$, we obtain
\[
  W_t\le(1-p_t)\varepsilon^2
      +p_t(1-p_t)\varepsilon^2
      =(1-p_t^2)\varepsilon^2
      \le\varepsilon^2.
\]
This proves the variance bound for all $t$. Finally, the sequence can switch only when an update occurs. Therefore, it holds that
\[
  \E[J(\vV_{1:T})]
  =\sum\nolimits_{t=2}^T\Pr(\vV_t\ne\vV_{t-1})
  \le\sum\nolimits_{t=2}^T p_t
  \le\frac{1}{\varepsilon}\sum\nolimits_{t=2}^T\Delta_t
  =\frac{P_T}{\varepsilon}.
\]

\subsection{Proof of Theorem~\ref{thm:sc_dynamic}}

Let $\gS(\vV_{1:T})$ denote the partition of $\vV_{1:T}$ into maximal constant intervals, so that $|\gS(\vV_{1:T})|-1=J(\vV_{1:T})$. Define its switching regret under the surrogate losses as
\begin{equation*}
    \SW_T^\ell(\vV_{1:T})
    =
    \sum\nolimits_{t=1}^T
    \bigl[\ell_t(\vy_t)-\ell_t(\vV_t)\bigr].
\end{equation*}
We state the switching-regret bound of IRESET-OGD for strongly convex losses \citep{ICML:2026:Yang}.
\begin{lemma}
  \label{lem:sc_ireset}
  Let $\Y$ be a convex domain contained in $B(\mathbf{0}, 4G/\lambda)$ and $\ell_1,\ldots,\ell_T$ be the surrogate losses in \eqref{eq:surrogate_str_cvx} with gradient norms at most $9G$. Then, IRESET-OGD ensures,  for every realization of $\vV_{1:T}$ in $\Y$ 
  \begin{equation}
    \label{eq:sc_ireset_bound}
    \SW_T^\ell (\vV_{1:T}) 
    \le \left[1+4(2+\log_2T)J(\vV_{1:T})\right]
        \left[2\Xi+\frac{81G^2}{\lambda}(1+\ln(2T))\right],
  \end{equation}
  where
  \begin{equation}
    \label{eq:sc_log_factors}
    \begin{aligned}
      \Xi
      &=144 \Gamma_T \frac{G^2}{\lambda}
        \left(2+\frac{1}{\sqrt{\ln2}}\right)
        +\frac{81 \Gamma_T^2G^2}{2\lambda\ln2},\\
      \Gamma_T
      &=3\ln2+\ln\left(1+\frac{1}{e}(1+\ln(2T+1))\right).
    \end{aligned}
  \end{equation}
\end{lemma}

Let $\Phi(T,J)$ denote the right-hand side of \eqref{eq:sc_ireset_bound}, with $J(\vV_{1:T})$ replaced by $J$. For any $0<\varepsilon\le2G/\lambda$, applying Theorem~\ref{thm:reduction} with this choice of $\Phi$ and $H=\lambda$, together with  $\E[J(\vV_{1:T})]\le P_T/\varepsilon$, yields

\begin{equation}
  \label{eq:sc_reduction_tradeoff}
  \begin{aligned}
\DReg_T(\vu_{1:T})
&\overset{\eqref{eq:f_l}}{\le}
\sum\nolimits_{t=1}^T
\bigl[\ell_t(\vy_t)-\ell_t(\vu_t)\bigr] \overset{\eqref{eq:reduction_bound}}{\le}
\E\bigl[\Phi(T,J(\vV_{1:T}))\bigr]
+\frac{\lambda T\varepsilon^2}{2}\\
    &\le \left(1+\frac{4(2+\log_2T)P_T}{\varepsilon}\right)
        \left[2\Xi +\frac{81G^2}{\lambda}(1+\ln(2T))\right]
        +\frac{\lambda T\varepsilon^2}{2},
  \end{aligned}
\end{equation}
of which the optimal minimizer is
\begin{equation*}
  \varepsilon_*=
  \left\{\frac{4(2+\log_2T)P_T}{\lambda T}
  \left[2\Xi +\frac{81G^2}{\lambda}(1+\ln(2T))\right]
  \right\}^{1/3}.
\end{equation*}
If $\varepsilon_*\le 2G/\lambda$, substituting it into \eqref{eq:sc_reduction_tradeoff} gives
\begin{equation}
  \label{eq:sc_unsubstituted_dynamic}
  \begin{aligned}
    \DReg_T(\vu_{1:T})
    &\le 2\Xi +\frac{81 G^2}{\lambda}(1+\ln(2T))\\
    &\quad+\frac32(\lambda T)^{1/3}
    \left\{4(2+\log_2T)P_T
    \left[2\Xi +\frac{81G^2}{\lambda}(1+\ln(2T))\right]
    \right\}^{2/3}.
  \end{aligned}
\end{equation}
If $\varepsilon_*>2G/\lambda$, according to $\diam(\X)\le2G/\lambda$, we have 
\begin{equation*}
  \DReg_T(\vu_{1:T}) \le G \sum\nolimits_{t=1}^T \|\vx_t - \vu_t\|_2 \le \frac{2G^2T}{\lambda} \le \lambda T\varepsilon_*^2,
\end{equation*}
where the last step is due to $\varepsilon_* \ge 2G/\lambda$, and $\lambda T\varepsilon_*^2$ is bounded by the right-hand side of \eqref{eq:sc_unsubstituted_dynamic}. Thus, \eqref{eq:sc_unsubstituted_dynamic} holds for every comparator sequence. Combining the two cases finishes the proof.

\section{Conclusion}
\label{sec:conclusion}

In this paper, we study dynamic regret minimization in non-stationary online learning and propose a novel framework that reduces this problem to switching regret minimization. With an unbiased random sequence and suitable surrogate losses, our framework allows us to use off-the-shelf switching-regret algorithms to establish the dynamic regret bounds of  $\widetilde{\gO}(T^{1/3}P_T^{2/3})$ for strongly convex losses and $\widetilde{\gO}(d^{2/3}T^{1/3}P_T^{2/3})$ for exp-concave losses. The same framework also ensures an $\gO(\sqrt{T(1+P_T)})$ dynamic regret bound for general convex losses. All our findings are  optimal for three types of losses, by a \textit{simple} and \textit{unified} analysis. Algorithmically, the auxiliary unbiased random sequence is used only for analysis, and need not be generated explicitly in practice. Our framework can thus use \textit{any} algorithm with switching-regret guarantees as a black box, without modifying its internal updates, to achieve corresponding dynamic regret bounds. 
\section*{AI Use Statement}

In this work, we used generative AI tools to polish the writing and assist with checking mathematical proofs. In particular, GPT-6-Astra helps us find a way to construct the auxiliary random sequence. We take full responsibility for the final content of this work, including text and claims developed with the assistance of generative AI.

\bibliography{ref}
\bibliographystyle{iclr2027_conference}

\clearpage
\appendix
\section{Proofs}
\label{app:dynamic_proofs}

In this part, we present the proofs of Theorems~\ref{thm:reduction}, \ref{thm:exp_dynamic}~and~\ref{thm:cov_dynamic}, followed by those of Lemma~\ref{lem:sc_ireset},~\ref{lem:exp_ireset}~and~\ref{lem:cov_reset}.

\subsection{Proof of Theorem~\ref{thm:reduction}}

We bound the two terms in \eqref{eq:dy:sl} separately. For the first term, applying the assumed switching-regret guarantee to each realization of $\vV_{1:T}$ and taking expectations gives
\begin{equation}
    \label{eq:reduction_switching_cost}
    \E\left[
        \sum\nolimits_{t=1}^T
        \bigl[\ell_t(\vy_t)-\ell_t(\vV_t)\bigr]
    \right]
    \le
    \E\bigl[\Phi(T,J(\vV_{1:T}))\bigr].
\end{equation}

For the second term, applying \eqref{eq:surrogate_loss} with $\vu=\vu_t$ and $\vv=\vV_t$ yields
\begin{equation}
    \label{eq:reduction_approximation_cost}
    \begin{aligned}
        \E\bigl[\ell_t(\vV_t)-\ell_t(\vu_t)\bigr]
        &\le
        \inner{\nabla\ell_t(\vu_t)}
              {\E[\vV_t-\vu_t]}
        +\frac{H}{2}
        \E\bigl[\norm{\vV_t-\vu_t}_2^2\bigr]\le \frac{H\varepsilon^2}{2},
    \end{aligned}
\end{equation}
where the last inequality follows from the unbiasedness and variance bound in Lemma~\ref{lem:unbiased_sequence}.

Summing \eqref{eq:reduction_approximation_cost} over all rounds and combining it with \eqref{eq:reduction_switching_cost} and \eqref{eq:dy:sl} gives \eqref{eq:reduction_bound}, completing the proof.
\subsection{Proof of Theorem~\ref{thm:exp_dynamic}}
\label{app:proof_exp}

We provide the switching regret bound of IRESET-ONS for exp-concave losses \citep{ICML:2026:Yang}.

\begin{lemma}
  \label{lem:exp_ireset}
  Let $\Y$ be a convex domain contained in $B(\mathbf{0},1/(16\beta G))$ and $\ell_1,\ldots,\ell_T$ be the surrogate losses in \eqref{eq:exp_surrogate} with gradient norms at most $\sqrt2G$. Then, IRESET-ONS ensures,  for every realization of $\vV_{1:T}$ in $\Y$ 
  \begin{equation}
    \label{eq:exp_ireset_bound}
    \begin{aligned}
      \SW_T^\ell(\vV_{1:T})
      &\le \left[1+4(2+\log_2T)J(\vV_{1:T})\right]  \left[2\Xi_{\mathrm{exp}}
        +\frac{5d}{\beta}\left(4+\frac{\sqrt2}{8}\right)(1+\ln(2T))\right],
    \end{aligned}
  \end{equation}
  where
  \begin{equation}
    \label{eq:exp_log_factors}
    \Xi_{\mathrm{exp}}
    =\frac{1}{\beta}\left[
      \frac{\sqrt2}{4}\Gamma_T\left(2+\frac{1}{\sqrt{\ln2}}\right)
      +\frac{4\Gamma_T^2}{\ln2}\right],
  \end{equation}
  with $\Gamma_T$ defined in \eqref{eq:sc_log_factors}.
\end{lemma}

For any $0<\varepsilon\le1/(32\beta G)$, combining Theorem~\ref{thm:reduction} with Lemma~\ref{lem:exp_ireset} and $\E[J(\vV_{1:T})]\le P_T/\varepsilon$ from Lemma~\ref{lem:unbiased_sequence} yields
\begin{equation}
  \label{eq:exp_reduction_tradeoff}
  \begin{aligned}
\DReg_T(\vu_{1:T})
&\overset{\eqref{eq:f_l}}{\le}
\sum\nolimits_{t=1}^T
\bigl[\ell_t(\vy_t)-\ell_t(\vu_t)\bigr] \overset{\eqref{eq:reduction_bound}}{\le}
\E\bigl[\Phi(T,J(\vV_{1:T}))\bigr]
+\frac{\beta G^2T\varepsilon^2}{2}\\
    &\le \left(1+\frac{4(2+\log_2T)P_T}{\varepsilon}\right)
      \left[2\Xi_{\mathrm{exp}}
        +\frac{5d}{\beta}\left(4+\frac{\sqrt2}{8}\right)(1+\ln(2T))\right]
      +\frac{\beta G^2T\varepsilon^2}{2},
  \end{aligned}
\end{equation}
of which the optimal minimizer is
\begin{equation}
  \label{eq:exp_solution}
  \varepsilon_*=
  \left\{\frac{4(2+\log_2T)P_T}{\beta G^2T}
    \left[2\Xi_{\mathrm{exp}}
      +\frac{5d}{\beta}\left(4+\frac{\sqrt2}{8}\right)(1+\ln(2T))\right]
  \right\}^{1/3}.
\end{equation}
If $\varepsilon_*\le1/(32\beta G)$, substituting it into \eqref{eq:exp_reduction_tradeoff} and enlarging the coefficient $3/2$ to $64$ gives
\begin{align}
  \DReg_T(\vu_{1:T})
  &\le 2\Xi_{\mathrm{exp}}
    +\frac{5d}{\beta}\left(4+\frac{\sqrt2}{8}\right)(1+\ln(2T))
    \label{eq:exp_explicit_dynamic}\\
  &\quad+64(\beta G^2T)^{1/3}
    \left[4(2+\log_2 T)P_T\right]^{2/3}
    \left[2\Xi_{\mathrm{exp}}
      +\frac{5d}{\beta}\left(4+\frac{\sqrt2}{8}\right)
      (1+\ln(2T))\right]^{2/3}.
    \nonumber
\end{align}
If $\varepsilon_*>1/(32\beta G)$, according to $\diam(\X)\le2R$ and $\beta GR\le1/32$, we have
\begin{equation*}
  \DReg_T(\vu_{1:T})
  \le G\sum\nolimits_{t=1}^T\norm{\vx_t-\vu_t}_2
  \le2GRT\le\frac{T}{16\beta}
  <64\beta G^2T\varepsilon_*^2,
\end{equation*}
where the last step is due to $\varepsilon_*>1/(32\beta G)$, and $64\beta G^2T\varepsilon_*^2$ is bounded by the right-hand side of \eqref{eq:exp_explicit_dynamic}. Thus, \eqref{eq:exp_explicit_dynamic} holds for every comparator sequence. Combining the two cases finishes the proof.

\subsection{Proof of Theorem~\ref{thm:cov_dynamic}}
\label{app:proof_cov}

We present the switching regret bound of RESET for general convex losses \citep{NeurIPS:2024:Pasteris}.

\begin{lemma}
\label{lem:cov_reset}
Let $\Y$ be a convex domain contained in $B(\mathbf{0},2R)$ and $\ell_1,\ldots,\ell_T$ be the surrogate losses in \eqref{eq:convex_surrogate} with gradient norms at most $G$. Then, RESET with OGD as its expert algorithm ensures, for every realization of $\vV_{1:T}$ in $\Y$,
\begin{equation}
\label{eq:cov_reset_bound}
  \SW_T^\ell(\vV_{1:T})
  \le 4\sqrt2 GR
  \left[\frac{\sqrt2}{\sqrt2-1}
       +\frac{\sqrt{8\ln2}}{3-2\sqrt2}\right]
       \sqrt{T\bigl(1+J(\vV_{1:T})\bigr)}.
\end{equation}
\end{lemma}

For any $0<\varepsilon\le R$, combining Theorem~\ref{thm:reduction} with Lemma~\ref{lem:cov_reset} and $\E[J(\vV_{1:T})]\le P_T/\varepsilon$ from Lemma~\ref{lem:unbiased_sequence} yields
\begin{equation*}
    \begin{aligned}
    \DReg_T(\vu_{1:T})
    &\overset{\eqref{eq:f_l}}{\le}
    \sum\nolimits_{t=1}^T
    \bigl[\ell_t(\vy_t)-\ell_t(\vu_t)\bigr] \overset{\eqref{eq:reduction_bound}}{\le} 
    \E\bigl[\Phi(T,J(\vV_{1:T}))\bigr]\\
      &\le 4\sqrt2 GR
      \left[\frac{\sqrt2}{\sqrt2-1}
           +\frac{\sqrt{8\ln2}}{3-2\sqrt2}\right]
           \sqrt{T\bigl(1+\E[J(\vV_{1:T})]\bigr)}\\
      &\le 4\sqrt2 GR
      \left[\frac{\sqrt2}{\sqrt2-1}
           +\frac{\sqrt{8\ln2}}{3-2\sqrt2}\right]
           \sqrt{T(1+P_T/\varepsilon)},
    \end{aligned}
\end{equation*}
where $H=0$ for the surrogate loss in \eqref{eq:convex_surrogate}, and the third inequality follows from Lemma~\ref{lem:cov_reset} and Jensen's inequality. Choosing $\varepsilon=R$ gives
\begin{equation*}
  \DReg_T(\vu_{1:T})
  \le 4\sqrt2\left[\frac{\sqrt2}{\sqrt2-1}
       +\frac{\sqrt{8\ln2}}{3-2\sqrt2}\right]
       G\sqrt{T(R^2+RP_T)}.
\end{equation*}

\subsection{Proof of Lemma~\ref{lem:sc_ireset}}
\label{app:proof_lemma6}

For a sequence $\vV_{1:T}$, let $\gS=\gS(\vV_{1:T})=\{\gI_1,\ldots,\gI_{|\gS|}\}$ be its partition set, so that $|\gS|-1=J(\vV_{1:T})$. We define $\SW_T^\ell(\gS)$ as in \eqref{eq:switching_regret_partition}, with $f_t$, $\vx_t$, and $\X$ replaced by $\ell_t$, $\vy_t$, and $\Y$, respectively. We apply the tree bound of \citet[Lemma~5]{ICML:2026:Yang} to the surrogate losses in \eqref{eq:surrogate_str_cvx}, and then consider $|\gS|=1$ and $|\gS|>1$ separately.

The surrogate has Hessian $\lambda I$ and gradient norms at most $9G$ on $\Y$, whose diameter is at most $8G/\lambda$. To handle arbitrary $T$, run IRESET-OGD with internal horizon $T^+=2^{\lceil\log_2T\rceil}<2T$. For the analysis, choose $\widehat{\vy}\in\arg\min_{\vy\in\Y}\sum_{t\in\gI_{|\gS|}}\ell_t(\vy)$ and extend the loss sequence by
\[
  \ell_t(\vy)=\frac{\lambda}{2}\norm{\vy-\widehat{\vy}}_2^2,
  \qquad T<t\le T^+.
\]
These additional losses are $\lambda$-strongly convex with gradient norms at most $8G\le9G$, so the source bound also applies to the extended sequence. Extend only the final interval to $T^+$ and denote the resulting partition by $\gS^+$. Since $\vV_{1:T}$ is constant on each interval of $\gS$, its cumulative loss on that interval is no smaller than that of the best fixed comparator. Thus, $\SW_T^\ell(\vV_{1:T})\le\SW_T^\ell(\gS)$.

We next compare the switching regret before and after extending the horizon. The added losses are nonnegative and vanish at $\widehat{\vy}$, a minimizer of the original cumulative loss on the final interval. Hence, extending this interval leaves its minimum cumulative loss unchanged. The other intervals are unaffected, and the learner's cumulative loss can only increase, since its predictions on $[T]$ remain unchanged. Consequently,
\[
  \SW_T^\ell(\vV_{1:T})
  \le \SW_T^\ell(\gS)
  \le \SW_{T^+}^\ell(\gS^+).
\]

On each dyadic interval $\gI$, the OGD expert with predictions $\vw_t$ and step sizes $1/[\lambda(t-\min\gI+1)]$ satisfies
\[
  \sum\nolimits_{t\in\gI}\ell_t(\vw_t)
  -\min_{\vy\in\Y}\sum\nolimits_{t\in\gI}\ell_t(\vy)
  \le\frac{81G^2}{\lambda}(1+\ln|\gI|).
\]
The initial term covers singleton intervals. Let $\mathcal F$ be the collection of maximal dyadic intervals contained in the intervals of $\gS^+$. Applying the global tree bound of \citet{ICML:2026:Yang} gives
\begin{equation}
\label{eq:sc_ireset_segments}
  \SW_T^\ell(\vV_{1:T})
  \le\SW_{T^+}^\ell(\gS^+)
  \le\sum\nolimits_{\gI\in\mathcal F}
  \left[2\Xi+\frac{81G^2}{\lambda}(1+\ln|\gI|)\right].
\end{equation}
Here the source horizon factor at $T^+$ is at most $\Gamma_T$, since $T^+<2T$. Substituting the gradient bound $9G$ and diameter bound $8G/\lambda$ into the source meta-regret constant gives exactly $\Xi$ in \eqref{eq:sc_log_factors}.

If $|\gS|=1$, the only maximal dyadic interval is the root $[T^+]$. Using $T^+<2T$ in \eqref{eq:sc_ireset_segments}, we obtain
\begin{equation*}
  \SW_T^\ell(\vV_{1:T})
  \le 2\Xi+\frac{81G^2}{\lambda}(1+\ln(2T)).
\end{equation*}

If $|\gS|>1$, each segment $\gI\in\gS^+$ contains at most two maximal dyadic intervals at each scale, hence at most $2(1+\lfloor\log_2|\gI|\rfloor)$ in total. Since every extended segment has length less than $2T$, $|\mathcal F|\le2|\gS|(2+\log_2T)$. Applying \eqref{eq:sc_ireset_segments} and $|\gS|\le2(|\gS|-1)$ gives
\begin{equation*}
  \begin{aligned}
    \SW_T^\ell(\vV_{1:T})
    &\le2|\gS|(2+\log_2T)
      \left[2\Xi+\frac{81G^2}{\lambda}(1+\ln(2T))\right]\\
    &\le4(|\gS|-1)(2+\log_2T)
      \left[2\Xi+\frac{81G^2}{\lambda}(1+\ln(2T))\right].
  \end{aligned}
\end{equation*}
Combining the two cases and using $|\gS|-1=J(\vV_{1:T})$ proves \eqref{eq:sc_ireset_bound}, which finishes the proof.

\subsection{Proof of Lemma~\ref{lem:exp_ireset}}
\label{app:proof_lemma7}

For a sequence $\vV_{1:T}$, let $\gS=\gS(\vV_{1:T})$ be its partition set, so $|\gS|-1=J(\vV_{1:T})$. We bound the switching regret using maximal dyadic intervals, treating $|\gS|=1$ and $|\gS|>1$ separately.

For the surrogate loss in \eqref{eq:exp_surrogate}, $\diam(\Y)\le1/(8\beta G)$ gives
\begin{equation*}
  \begin{aligned}
    \left|\beta\inner{\vh_t}{\vy-\vy_t}\right|&\le\frac18,\\
    \norm{\nabla\ell_t(\vy)}_2
    &=\left|1+\beta\inner{\vh_t}{\vy-\vy_t}\right|\norm{\vh_t}_2
      \le\frac98G\le\sqrt2G.
  \end{aligned}
\end{equation*}
Together with $\nabla^2\ell_t(\vy)=\beta\vh_t\vh_t^\top$, this gives $\nabla^2\ell_t(\vy)\succeq(\beta/4)\nabla\ell_t(\vy)\nabla\ell_t(\vy)^\top$, so $\ell_t$ is $\beta/4$-exp-concave. The exp-concavity inequality of \citet[Lemma~1]{ICML:2026:Yang} therefore gives
\begin{equation*}
  \ell_t(\vv)\ge\ell_t(\vu)+\inner{\nabla\ell_t(\vu)}{\vv-\vu}
    +\frac{\beta}{16}\inner{\nabla\ell_t(\vu)}{\vv-\vu}^2,
  \qquad \vu,\vv\in\Y,
\end{equation*}
since $\min\{\beta/8,\beta/\sqrt2\}=\beta/8$. Using this coefficient, the gradient bound $\sqrt2G$, and the diameter bound $1/(8\beta G)$ in the meta-regret calculation of \citet{ICML:2026:Yang} gives $\Xi_{\mathrm{exp}}$ in \eqref{eq:exp_log_factors}.

Initialize each ONS expert at a point in $\Y$ with $A_0=(64G)^2I$ and step size $8/\beta$. On an interval $\gI$, let $\vw_t$ denote the expert's prediction at round $t\in\gI$. The ONS analysis \citep{Others:2007:Hazan}, with the initialization term retained, gives, for every $\vu\in\Y$,
\begin{equation}
  \label{eq:lemma7_ons}
  \begin{aligned}
    \sum\nolimits_{t\in\gI}\bigl[\ell_t(\vw_t)-\ell_t(\vu)\bigr]
    &\le\frac4\beta\left[d\ln\left(1+\frac{|\gI|}{2048}\right)+1\right] 
     \le\frac{5d}{\beta}\left(4+\frac{\sqrt2}{8}\right)(1+\ln|\gI|),
  \end{aligned}
\end{equation}
which follows from $d\ge1$ and $\ln(1+|\gI|/2048)\le\ln|\gI|+1/2048$, and includes singleton intervals.

Run IRESET-ONS with internal horizon $T^+=2^{\lceil\log_2T\rceil}<2T$. For the analysis, append zero losses, obtained from \eqref{eq:exp_surrogate} by setting $\vh_t=\mathbf0$ for $t>T$, and extend only the final segment of $\gS$ to obtain $\gS^+$. This continuation preserves the predictions on $[T]$, the cumulative losses, and the number of segments, so
\begin{equation*}
  \SW_T^\ell(\vV_{1:T})\le\SW_T^\ell(\gS)=\SW_{T^+}^\ell(\gS^+).
\end{equation*}
The source horizon factor at $T^+$ is bounded by $\Gamma_T$ because $T^+<2T$.

Let $\mathcal F$ be the partition into maximal dyadic intervals contained in the segments of $\gS^+$, using the complete binary tree on $[T^+]$. Applying the recursion of \citet[Lemmas~8--9]{ICML:2026:Yang} to $\gS^+$, together with \eqref{eq:lemma7_ons}, yields
\begin{equation}
  \label{eq:lemma7_dyadic}
  \begin{aligned}
    \SW_T^\ell(\vV_{1:T})
    &\le\sum\nolimits_{\gI\in\mathcal F}
      \left[2\Xi_{\mathrm{exp}}
        +\frac{5d}{\beta}\left(4+\frac{\sqrt2}{8}\right)(1+\ln|\gI|)\right].
  \end{aligned}
\end{equation}

If $|\gS|=1$, the only maximal dyadic interval is the root $[T^+]$. Using $T^+<2T$ in \eqref{eq:lemma7_dyadic}, we obtain
\begin{equation*}
  \SW_T^\ell(\vV_{1:T})
  \le2\Xi_{\mathrm{exp}}
    +\frac{5d}{\beta}\left(4+\frac{\sqrt2}{8}\right)(1+\ln(2T)).
\end{equation*}

If $|\gS|>1$, each segment $\gI\in\gS^+$ contains at most two maximal dyadic intervals at each scale, hence at most $2(1+\lfloor\log_2|\gI|\rfloor)$ in total. Since every extended segment has length less than $2T$, $|\mathcal F|\le2|\gS|(2+\log_2T)$. Applying \eqref{eq:lemma7_dyadic} and $|\gS|\le2(|\gS|-1)$ gives
\begin{equation*}
  \begin{aligned}
    \SW_T^\ell(\vV_{1:T})
    &\le4(|\gS|-1)(2+\log_2T)
     \left[2\Xi_{\mathrm{exp}}
      +\frac{5d}{\beta}\left(4+\frac{\sqrt2}{8}\right)(1+\ln(2T))\right].
  \end{aligned}
\end{equation*}
Combining the two cases and using $|\gS|-1=J(\vV_{1:T})$ proves \eqref{eq:exp_ireset_bound}.
\subsection{Proof of Lemma~\ref{lem:cov_reset}}
\label{app:proof_lemma8}

For a sequence $\vV_{1:T}$, let $\gS=\gS(\vV_{1:T})$ be its partition set, so that $|\gS|-1=J(\vV_{1:T})$. We apply \citet[Theorem~2.2]{NeurIPS:2024:Pasteris} to the normalized surrogate losses in \eqref{eq:convex_surrogate}.
For the surrogate in \eqref{eq:convex_surrogate}, the normalized losses are
\begin{equation*}
  \widetilde\ell_t(\vy)
  =\frac{\inner{\vh_t}{\vy}+2R\norm{\vh_t}_2}{4GR}
  \in[0,1],
  \qquad \vy\in\Y,
\end{equation*}
whose gradient norms are at most $1/(4R)$, and normalization divides regret by
$4GR$.
For an OGD expert on an interval $\gI$, use step size
$\eta_{\gI}=16R^2/\sqrt{|\gI|}$ and projection onto $\Y$.
Let $\vw_t$ denote this expert's prediction at round $t\in\gI$.
Since $\diam(\Y)\le4R$, starting from any point in $\Y$ gives
\[
  \sum\nolimits_{t\in\gI}\widetilde\ell_t(\vw_t)
  -\min_{\vu\in\Y}\sum\nolimits_{t\in\gI}\widetilde\ell_t(\vu)
  \le\frac{16R^2}{2\eta_{\gI}}
      +\frac{\eta_{\gI}|\gI|}{32R^2}
  =\sqrt{|\gI|}.
\]
Thus the source theorem applies with expert regret constant $\gamma=1$,
including $|\gI|=1$.

For an arbitrary horizon $T$, use the internal horizon
$T^+=2^{\lceil\log_2T\rceil}<2T$ and append zero losses after round $T$.
Extend only the final interval of $\gS$ to $T^+$ and denote the resulting partition
by $\gS^+$, so that $|\gS^+|=|\gS|$ and
$\sum_{\gI\in\gS^+}|\gI|=T^+$.
The continuation leaves all predictions on $[T]$ unchanged, and zero losses do
not change the learner's cumulative loss or any interval's minimum cumulative loss.
Applying Theorem~2.2 on $[T^+]$ therefore gives
\begin{equation}
\label{eq:cov_reset_theorem}
\begin{aligned}
  \SW_T^\ell(\vV_{1:T})
  \le 4GR\,\SW_{T^+}^{\widetilde\ell}(\gS^+) 
   \le 4GR
  \left[\frac{\sqrt2}{\sqrt2-1}
       +\frac{\sqrt{8\ln2}}{3-2\sqrt2}\right]
       \sum\nolimits_{\gI\in\gS^+}\sqrt{|\gI|}.
\end{aligned}
\end{equation}
Finally, Cauchy--Schwarz and $T^+<2T$ yield
\[
  \sum\nolimits_{\gI\in\gS^+}\sqrt{|\gI|}
  \le\sqrt{|\gS|\sum\nolimits_{\gI\in\gS^+}|\gI|}
  \le\sqrt{2T\bigl(1+J(\vV_{1:T})\bigr)}.
\]
Substituting this inequality into \eqref{eq:cov_reset_theorem} proves
\eqref{eq:cov_reset_bound}, which finishes the proof.

\end{document}